\documentclass{article} 
\usepackage{iclr2025_conference,times}

\usepackage{amsmath,amsfonts,bm}

\def\eqref#1{equation~\ref{#1}}

\def\1{\bm{1}}

\DeclareMathAlphabet{\mathsfit}{\encodingdefault}{\sfdefault}{m}{sl}
\SetMathAlphabet{\mathsfit}{bold}{\encodingdefault}{\sfdefault}{bx}{n}

\usepackage[hyperfootnotes=false]{hyperref}
\usepackage{url}
\usepackage{graphicx}
\usepackage{booktabs} 
\usepackage{multirow} 
\usepackage{siunitx}  
\usepackage[inkscapelatex=false]{svg}
\usepackage{ragged2e}
\usepackage[dvipsnames]{xcolor}
\usepackage[table]{xcolor}
\usepackage{marvosym}
\usepackage{algorithm}
\usepackage{algpseudocode}
\usepackage[most]{tcolorbox}
\usepackage{booktabs,array,multirow,threeparttable,xcolor}
\usepackage{pifont}
\usepackage{siunitx}
\usepackage{booktabs}
\usepackage{multirow}
\usepackage{array}
\usepackage{float}
\usepackage{graphicx}

\definecolor{mygray}{RGB}{220, 220, 220}
\newcommand\blfootnote[1]{%
  \begingroup
  \renewcommand\thefootnote{}\footnote{#1}%
  \addtocounter{footnote}{-1}%
  \endgroup
}

\title{MHPR: Multidimensional Human Perception and Reasoning Benchmark for Large Vision-Languate Models}

\author{Kangkang Wang$^{1, \textrm{\Letter},\textrm{\Cross}}$, Qinting Jiang$^{2, \textrm{\Cross}}$, Wanping Zhang$^{1}$, Bowen Ren$^{1}$, Shengzhao Wen$^{1}$\\
$^1$AI Compute Group, Baidu $\quad ^2$Tsinghua University \\
}

\iclrfinalcopy 
\begin{document}

\maketitle

\blfootnote{
$\dagger$ Equal Contribution\\
\Letter~Emails: wangkangkang@baidu.com
}

\begin{abstract}
    Multidimensional human understanding is essential for real-world applications such as film analysis and virtual digital humans, yet current LVLM benchmarks largely focus on single-task settings and lack fine-grained, human-centric evaluation. In this work, we introduce MHPR, a comprehensive benchmark for joint perception–reasoning over human-centric scenes spanning individual, multi-person, and human–object interaction dimensions. MHPR comprises a multi-level data design—Captioned Raw Data (C-RD), Supervised Fine-Tuning Data (SFT-D), Reinforcement Learning Data (RL-D), and Test Data (T-D)—together with an automated caption/VQA generation pipeline (ACVG) that performs category-wise attribute decomposition, attribute-specific rewriting, and multi-model voting to ensure high-quality, scalable annotations. We evaluate state-of-the-art vision–language models on fine-grained attributes (appearance, clothing, pose, parts) and high-level semantics (social relations, action semantics, spatial relations, intent and functionality). Our findings show that: 1) format-aligned SFT data substantially improves instruction following and stability; 2) challenge-focused RL data derived from bad-case analysis further enhances perception and reasoning on difficult instances; and 3) training Qwen2.5-VL-7B with MHPR yields significant gains, achieving near-parity with considerably larger models. We release ACVG and MHPR to facilitate reproducible, extensible research on human-centric perception and reasoning. 
\end{abstract} 
\section{Introduction}

In recent years, with the rapid development of intelligent retail, virtual digital humans, and film/video content analysis, automated and intelligent human-centric understanding has become a key technical foundation \cite{fu2025video, li2024mvbench, zhouhumanvbench, dong2025moga, chen2025socialnav}. The requirements in these real-world scenarios go far beyond simple recognition and detection of “people.” They demand fine-grained perception of an individual’s appearance, clothing, actions, and identity, as well as an understanding of spatial and social relations among multiple people and complex human–object interactions, emotions, and functionality. For example, in film content analysis, accurately modeling group relationships, character identities, and dynamic interactions is central to intelligent content processing. However, existing datasets and Large Vision-Language Models (LVLMs) face severe limitations in supporting these needs, forming a bottleneck for downstream application intelligence \cite{yu2016modeling, jiang2025referring, qin2025face}.

Recent works such as RefCOCO \cite{yu2016modeling} and HumanRef \cite{jiang2025referring} have proposed human-centric detection benchmarks to evaluate and improve language-guided target localization (especially for human targets). Building on this line, HERM-Bench \cite{li2024herm}  further introduces simple VQA tasks. Nevertheless, these efforts largely focus on single-task dimensions (e.g., detection or classification) and lack fine-grained annotations across the multidimensional spectrum of “individual–multi-person–human–object.” As a result, they fall short of supporting comprehensive understanding of human–human and human–object interactions in complex real-world scenes. This leads to multiple practical limitations: in virtual digital human generation, the lack of multidimensional, fine-grained human and interaction data results in limited visual diversity, less realistic social behavior, and weak semantic consistency between dialogue and actions; in film content analysis, systems struggle to robustly recognize group relations, character identities, and dynamic interactions within and across shots and narrative threads, thereby hindering character tracking, summarization, and content retrieval. The main differences between MHPR and other benchmarks are summarized in Table \ref{tab:Benchmarks}.

\begin{table*}[t]
\centering
\small
\begin{threeparttable}
\setlength{\tabcolsep}{8pt}
\renewcommand{\arraystretch}{1.2}
\newcommand{\cmark}{\ding{51}}
\newcommand{\xmark}{\ding{55}}

\begin{tabular}{
    l
    |
    c
    c
    c
    c
    S[table-format=2.1,table-space-text-post= k,table-number-alignment=center]
}
\toprule
& \textbf{V-Grouding} & \textbf{R-Grounding} & \textbf{V-Perception} & \textbf{V-Reasoning} & {\# of QA} \\
\midrule
RefCoCo      & \cmark & \xmark & \cmark & \xmark & 41k  \\
HERM-Bench         & \cmark & \xmark & \cmark & \xmark & 33.4k\\
HC-RefLoCo      & \cmark & \cmark & \xmark & \xmark & 720k \\
HumanRef         & \cmark & \xmark & \cmark & \xmark & 10.9k\\
Face-Human  & \cmark & \xmark & \cmark & \xmark & 50.4k\\
\midrule
MHPR (Ours)                    & \cmark & \cmark & \cmark & \cmark & 9.0k \\
\bottomrule
\end{tabular}
\end{threeparttable}
\caption{Comparison across benchmarks RefCoCo \cite{yu2016modeling}, HERM-Bench \cite{li2024herm}, HC-RefLoCo \cite{wei2024large}, HumanRef \cite{jiang2025referring}, Face-Human \cite{qin2025face}  and MHPR (Ours). V-Grouding: Visual Grouding; R-Grounding: Reference Grouding; V-Perception: Visual perception; V-Reasoning: Visual Reasoning.}
\label{tab:Benchmarks}
\end{table*}

To bridge this gap, we present a systematic and comprehensive human-centric benchmark: Multidimensional Human Perception and Reasoning (MHPR). MHPR goes beyond basic capabilities such as person identification and localization, and conducts a joint evaluation along two core dimensions: perception and reasoning. Concretely, we organize evaluation across three complementary scenario dimensions—individual, multi-person, and human–object—to systematically capture individual attributes, group relations, and interaction mechanisms. On the perception side, MHPR requires fine-grained human understanding, including appearance, clothing, accessories, pose, and local attributes. On the reasoning side, it further assesses the understanding of social attributes, action semantics, spatial relations, and the intent and functionality underlying human–human and human–object interactions. This design moves models beyond the mechanical verification of a target’s presence toward deeper, context-aware comprehension of human motives, social relations, and situational interactions, thereby aligning with the high-level semantic and complex reasoning requirements in applications such as film analysis and virtual digital humans.

Within MHPR, we introduce two key components: (1) a multi-level data structure and (2) an automated captioning and VQA generation pipeline, ACVG. The multi-level data structure comprises four data types—Captioned Raw Data (C-RD), Supervised Fine-Tuning Data (SFT-D), Reinforcement Learning Data (RL-D), and Test Data (T-D)—with clear division of roles and seamless interoperation:
\begin{itemize}
\item C-RD (Captioned Raw Data): preserves the openness and extensibility of the benchmark. Researchers can use C-RD to customize new tasks and evaluation dimensions in future work, supporting continual evolution and domain transfer.
\item SFT-D (Supervised Fine-Tuning Data): consists of large-scale, task-related VQA samples whose elements and format are strictly aligned with T-D. It is used to learn fundamental instruction-following abilities, reinforce pattern imitation and knowledge transfer, and provide a “usable and stable” initial policy and alignment baseline.
\item RL-D (Reinforcement Learning Data): serves as an advanced set built upon SFT-D. We conduct systematic bad-case analysis on model errors in T-D, summarize the main challenges posed by MHPR tasks, and construct targeted VQA samples to specifically enhance perception and reasoning capabilities.
\item T-D (Test Data): used for objective evaluation and kept consistent with SFT-D in composition and annotation paradigm, ensuring training–evaluation consistency and comparability.
\end{itemize}
For data generation, prior work often adopts a “model generation + manual curation” pipeline, which is labor-intensive and poorly scalable. ACVG is the first to introduce a multi-model voting mechanism, iteratively correcting inconsistencies in generated captions and VQA items to substantially improve annotation quality and semantic consistency. With minimal reliance on manual curation, ACVG produces high-quality data at scale, significantly reducing labor costs and improving throughput. The combination of a “multi-level data structure + automated ACVG pipeline” ensures both extensibility and task alignment, while providing systematic data support for the SFT→RL capability progression.

We train Qwen2.5-VL-7B using MHPR. Experimental results show substantial performance gains across multiple metrics, achieving near-parity with considerably larger models. Our main contributions are summarized as follows:
\begin{itemize}
\item We present MHPR, a human-centric, multidimensional benchmark. MHPR jointly evaluates perception and reasoning across three scenario dimensions—individual, multi-person, and human–object interaction—covering fine-grained attributes (appearance, clothing, pose, parts) and high-level semantics (social relations, action semantics, spatial relations, intent and functionality), thereby addressing key gaps in complex human-scene understanding.
\item We introduce a multi-level data design and an automated caption/VQA generation pipeline, ACVG. The data stack comprises C-RD, SFT-D, RL-D, and T-D: C-RD preserves openness and extensibility; SFT-D is format-aligned with T-D to provide a stable instruction-following baseline; RL-D is constructed via bad-case analysis on T-D to target core challenges with advanced VQA for perception and reasoning. ACVG employs multi-model voting, consistency correction, and targeted attribute rewriting, enabling selective correction of designated attributes while preserving the remaining content of the original caption as much as possible, thereby producing high-quality captions and VQA with minimal human curation, greatly reducing manual costs and improving scalability.
\item Using MHPR, we train Qwen2.5-VL-7B and observe substantial improvements across multiple metrics, achieving near-parity with much larger models. These results validate MHPR’s effectiveness in enhancing fine-grained perception and high-level reasoning for human-centric scenarios.
\end{itemize}
\section{MHPR Benchmark}
\subsection{Data Source Selection}
Our data sources consist of three open-source datasets from different domains that provide semantically rich raw corpora: COYO-700M contributes a large-scale, open-domain, stylistically diverse collection of image–text pairs; HumanCaption-10M offers broad-coverage, fine-grained, human-centric descriptions; and HumanCaption-HQ-311K serves as a high-confidence, low-noise refined subset for alignment and calibration. Together, they complement each other to provide breadth, depth, and stability for MHPR’s data generation, task alignment, and evaluation:
\begin{itemize}
\item \textbf{HumanCaption-HQ-311K} \cite{dai2025humanvlm} is an open-source, high-quality human-centric caption dataset designed to train and evaluate multimodal/vision-language models on fine-grained human and scene understanding. It contains approximately 311K high-quality image captions with auxiliary metadata, covering multi-dimensional aspects such as appearance, clothing, pose, actions, local attributes, and contextual human–human/human–object interactions. The data are produced via a semi-automated pipeline with multi-model consistency checking, followed by rigorous quality control and denoising to balance coverage and accuracy. HumanCaption-HQ-311K supports supervised fine-tuning, data augmentation, and instruction alignment, and is particularly suitable for human-centric fine-grained perception and reasoning tasks, including person retrieval, action understanding, relation recognition, and human–object interaction comprehension.
\item \textbf{HumanCaption-10M} \cite{dai2025humanvlm} is a large-scale, open-source human-centric caption dataset designed to supply broad-coverage, fine-grained, and context-rich training corpora for multimodal/vision-language models. It comprises approximately 10 million image captions with accompanying metadata, spanning appearance, detailed clothing attributes, pose and actions, facial expressions and local parts, as well as multi-person interactions, gaze and spatial relations, and contextual human–object manipulation and functional cues. Built via an efficient semi-automated pipeline with multi-model consistency checking, noise suppression, and quality stratification, HumanCaption-10M balances scale and accuracy, supporting supervised fine-tuning, instruction alignment, and data augmentation. It is particularly suitable for enhancing generalization and robustness on tasks such as person retrieval, action and relation understanding, human–object interaction, fine-grained attribute recognition, and narrative consistency modeling.
\item \textbf{COYO-700M} \cite{lu2023delvingdeeperdatascaling} is a large-scale, open-source image–text pair dataset designed as a general-purpose pretraining corpus for multimodal/vision-language models. It comprises approximately 700 million image–text pairs collected from diverse web sources, spanning a wide range of domains, visual styles, and linguistic expressions, with strong diversity and long-tail coverage. The dataset is preprocessed via large-scale deduplication, basic safety and quality filtering, text cleaning, and language identification to improve usability and robustness at scale. COYO-700M supports contrastive pretraining (CLIP-style), image–text matching, cross-modal retrieval, general CPT/pretraining, and serves as a broad-coverage foundation for downstream SFT, substantially enhancing cross-domain generalization.
\end{itemize}

\subsection{Data Dimension}



    



\begin{table}[H] 
\centering
\normalsize
\setlength{\tabcolsep}{2.6pt}      
\renewcommand{\arraystretch}{1.04} 
\newcolumntype{C}[1]{>{\centering\arraybackslash}m{#1}}

\begin{tabular}{C{2.3cm}|C{2.5cm} C{0.64\linewidth}}
\toprule
\textbf{Category} & \textbf{Topic} & \textbf{Description} \\
\midrule
\multirow{2}{*}{Single Per.}
  & Appearances & Hairstyle, facial features, body build, etc.\\
  & Attributes  & Age, gender, ethnicity, occupation, etc.\\
\midrule
\multirow{2}{*}{Multiple Per.}
  & Relations    & Relations between multiple individuals. \\
  & Comparisons  & Commonalities or differences between individuals. \\
\midrule
\multirow{1}{*}{Per. and Obj.}
  & Interactions & Interactions between a target person and objects. \\
\bottomrule
\end{tabular}
\caption{Our evaluation dimensions primarily include Single Person, Multiple Persons, and Person–Object Interaction, covering multiple levels from perception to reasoning.}
\label{tab:dimensions}
\end{table}

After determining the data sources, we annotate captions and construct VQA tasks for the raw data. We divide the data into three dimensions—single person, multiple persons, and person–object—capturing multidimensional interactions between humans and themselves, society, and the surrounding environment. Specifically, each dimension includes the following attributes:
\begin{itemize}
    \item \textbf{Single person}. Appearance: hairstyle, facial features, body shape, skin tone, clothing, accessories, pose and actions, expressions, etc.; Attributes: age, gender, ethnicity, occupation; Referential localization: identify a specific person based on explicit attributes such as appearance and pose.

    \item \textbf{Multiple persons}. Interpersonal relations: understand various relationships among multiple individuals in the image, such as interactions, social ties, and spatial relations; Comparative analysis: analyze similarities and differences between individuals, e.g., commonalities/differences in attire and identity.

    \item \textbf{Person and object}. Identify and reason about interactions between a specific person and non-human objects in the image.
\end{itemize}

With this partitioning, we extend beyond standalone human recognition to encompass human–environment interactions, mirroring diverse states in the real world. In the subsequent caption annotation and VQA generation, we will address these three dimensions separately. The details can be seen in Table \ref{tab:dimensions}.

\subsection{Automated Captioning and VQA Generation Pipeline (ACVG)}
In ACVG, we aim to leverage existing LVLMs to automatically produce high-quality caption annotations and VQA tasks, thereby improving labeling efficiency and reducing manual screening costs. ACVG comprises two components: a caption auto-annotation pipeline and a VQA auto-generation pipeline.

\subsubsection{Caption Auto-Annotation Pipeline (CAAP)}


In existing work, images first undergo a preliminary screening to filter out low-quality samples. The filtered samples are then captioned once by an LVLM (e.g., GPT‑4o). These captioned images are subsequently reviewed by human annotators, which becomes labor-intensive at scale. Our pipeline, CAAP, focuses on two goals: (1) automating the filtering of low-quality samples, and (2) automatically producing accurate captions for the screened images.

\textbf{Pipline for automating the filtering of low-quality samples.} To ensure reliable downstream human-attribute analysis and VQA construction, we build a four-stage screening pipeline: human detection, scale filtering, keypoint completeness checking, and collage-image filtering. We rely on reproducible open-source implementations and calibrate key thresholds on a development split. The specific implementation details are as follows:

\paragraph{Stage 1: Human Detection and Initial Filtering}
We adopt the Group-DINO human detector from PaddleDetection with the configuration \texttt{configs/group\_detr/group\_dino\_vit\_huge\_4scale\_1x\_coco.yml}. The detection confidence threshold is set to $0.5$, selected after scanning $[0.3, 0.7]$ to balance recall and false positives. Images with no human bounding boxes meeting the threshold are discarded.

\paragraph{Stage 2: Human Scale Filtering}
For each image, we compute the ratio $r = A_{\max}/A_{\text{img}}$, where $A_{\max}$ is the area of the largest human bounding box and $A_{\text{img}}$ is the image area. If $r < 0.10$, the image is removed. This eliminates cases where the main subject is too small to support reliable attribute extraction (e.g., apparel texture, facial expression, accessories). 

\paragraph{Stage 3: Keypoint Completeness Check}
We employ MMPose YOLOX-Pose with configuration \texttt{configs/yolox-pose\_s\_8xb32-300e\_coco.py}. Human detections from Stage~1 are used to crop/align regions, on which we run keypoint detection. For each human box, we count keypoints with confidence above $0.5$; if fewer than $4$ keypoints are present, the instance is considered incomplete (occlusion, truncation, severe blur) and is discarded. If no instances in an image satisfy the rule, the entire image is removed.

\paragraph{Stage 4: Collage-Image Filtering}
Web-crawled data often contains collages (multi-panel grids, comparison plates, product mosaics), which differ markedly from natural scenes and hinder attribute modeling. We curated a binary classification dataset of $\sim$20k images (collage vs. non-collage) with manual labels, and trained a classifier using PaddleClas ResNet50\_vd (\texttt{./ppcls/configs/quick\_start/ResNet50\_vd.yaml}). Training uses standard augmentations (random crop, horizontal flip), cosine-annealed learning rate, early stopping, and class reweighting or resampling if imbalanced. At inference, images with $P(\text{collage}) \ge \tau$ (default $\tau=0.5$) are discarded. The complete procedure can be found in Algorithm \ref{algorithm: HCDC}.

\begin{algorithm}[t]
\caption{Human-centric Data Cleaning Pipeline}
\label{algorithm: HCDC}
\begin{algorithmic}[1]
\Require Image set $\mathcal{I}$; detectors $\mathcal{D}_{\text{human}}$, $\mathcal{D}_{\text{kpt}}$; collage classifier $\mathcal{C}$; thresholds $\theta=0.5$ (bbox score), $\rho=0.10$ (area ratio), $\kappa=0.5$ (kpt score), $K_{\min}=4$ (min keypoints), $\tau=0.5$ (collage prob).
\Ensure Filtered image set $\mathcal{I}^\star$.
\State $\mathcal{I}^\star \leftarrow \emptyset$.
\ForAll{$I \in \mathcal{I}$}
  \State $\mathcal{B} \leftarrow \{b \mid \mathcal{D}_{\text{human}}(I) \text{ with } \text{score}(b)\ge \theta\}$
  \If{$\mathcal{B}=\emptyset$}
    \State \textbf{continue} \Comment{discard: no human}
  \EndIf
  \State Compute $r = \max_{b\in\mathcal{B}} \frac{\text{area}(b)}{\text{area}(I)}$
  \If{$r < \rho$}
    \State \textbf{continue} \Comment{discard: subject too small}
  \EndIf
  \State $\mathcal{B}' \leftarrow \emptyset$
  \ForAll{$b \in \mathcal{B}$}
    \State Crop $I$ to region $I_b$ aligned with $b$
    \State $\mathcal{K}_b \leftarrow \{k \mid \mathcal{D}_{\text{kpt}}(I_b),\ \text{score}(k)\ge \kappa\}$
    \If{$|\mathcal{K}_b| \ge K_{\min}$}
      \State $\mathcal{B}' \leftarrow \mathcal{B}' \cup \{b\}$
    \EndIf
  \EndFor
  \If{$\mathcal{B}'=\emptyset$}
    \State \textbf{continue} \Comment{discard: incomplete humans}
  \EndIf
  \State $p \leftarrow \mathcal{C}(I)$ \Comment{probability of collage}
  \If{$p \ge \tau$}
    \State \textbf{continue} \Comment{discard: collage image}
  \EndIf
  \State $\mathcal{I}^\star \leftarrow \mathcal{I}^\star \cup \{I\}$
\EndFor
\State \Return $\mathcal{I}^\star$
\end{algorithmic}
\end{algorithm}

\textbf{Pipline for automatically producing accurate captions for the screened images.} This pipline consolidates multiple locally generated captions into a single, more accurate human-centric description, minimizing the need for manual review. The system comprises three stages: (i) structured disagreement detection (GetDiff), (ii) conflict correction via multi-source voting (Vote), and (iii) schema-level fusion into a final caption (Mix).

\begin{figure}[htbp] 
\centering
\includegraphics[width=\textwidth]{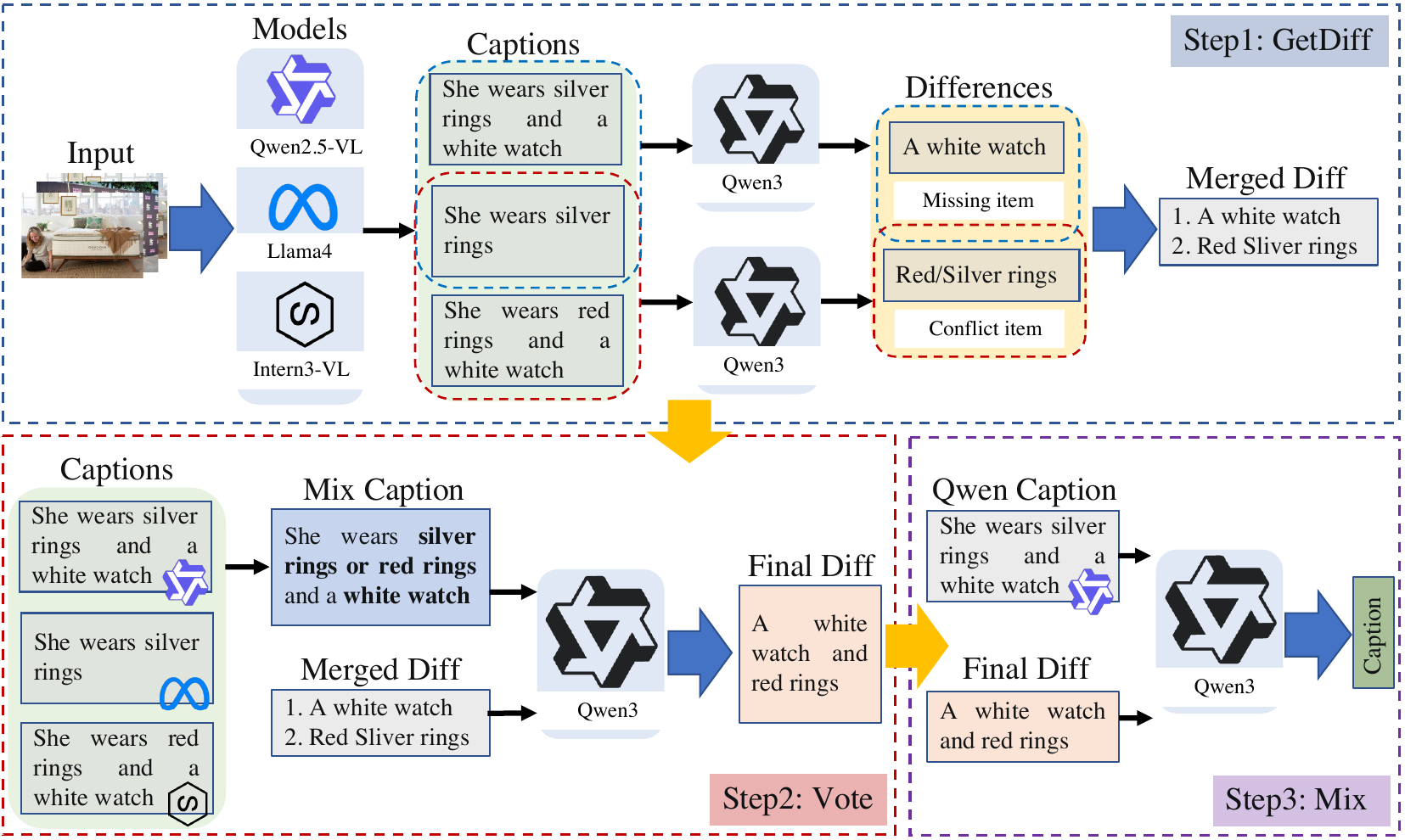}
\caption{Overall pipeline for \textbf{CAAP}, including three steps: GetDiff, Vote and Mix.}
\label{fig:CAAP_pipline}
\end{figure}

We define a fixed schema for human-related attributes (e.g., gender, age group, apparel categories and colors, accessories, actions, and scene context). Each attribute has required fields and admissible values. All intermediate artifacts are stored as JSON dictionaries keyed by attribute, enabling deterministic merging and auditability. The details can be found in Figure \ref{fig:CAAP_pipline}.

\paragraph{Step 1: GetDiff — Structured Disagreement Detection.}

\textbf{Inputs.} Three local captions describing the same image, produced by \textit{qwen}, \textit{intern} and \textit{llama}.

\noindent\textbf{Procedure.}
\begin{enumerate}
  \item A strong LLM (qwen3-32B in our implementation) parses each caption into the predefined attribute schema.
  \item For every attribute, the parsed fields are compared and labeled as one of three categories:
  \begin{itemize}
    \item \textit{Conflict}: both captions mention the attribute but provide incompatible values;
    \item \textit{Missing}: the attribute appears in only one caption;
    \item \textit{No-conflict}: values are identical or semantically equivalent.
  \end{itemize}
\end{enumerate}

\noindent\textbf{Outputs.} A \textit{diff dictionary} listing all conflicted or missing attributes together with evidence snippets from each caption. Because the output is structured, iterating over the dictionary directly reveals which attributes require correction.

\paragraph{Step 2: Vote — Conflict Correction with Multi-Caption Evidence.}
\textbf{Inputs.} The diff dictionary and three local captions (\textit{qwen}, \textit{intern}, \textit{llama4}) as reference evidence.

\noindent\textbf{Procedure.}
\begin{enumerate}
  \item For each conflicted or missing attribute, aggregate the corresponding fields from all three captions and present them (together with the image) to the LLM.
  \item Instruct the LLM to ``vote'': prefer majority agreement when present; when disagreement persists, defer to visual grounding cues.
  \item The LLM returns a corrected value with a confidence score and an optional brief rationale for auditing.
\end{enumerate}

\noindent\textbf{Outputs.} A \textit{correction dictionary} containing the resolved values for all conflicted/missing attributes.

\paragraph{Step 3: Mix — Fusion into the Final Caption.}
\textbf{Inputs.} The baseline \textit{qwen} local caption and the correction dictionary.

\noindent\textbf{Procedure.}
\begin{enumerate}
  \item Parse the baseline caption into the attribute schema.
  \item Overwrite conflicted/missing fields using the correction dictionary.
  \item Ask the LLM to realize the updated schema as a fluent, image-grounded caption while ensuring internal consistency.
\end{enumerate}

\noindent\textbf{Outputs.} A single, more accurate final caption per image.

\paragraph{Implementation Details.}
\begin{itemize}
  \item \textbf{Caption generators:} qwen2.5-vl-72B, intern3-vl-78B, and llama-4-maverick-17B-128e-instruct produce the local captions.
  \item \textbf{Orchestrator:} qwen3-32B executes \textit{GetDiff}, \textit{Vote}, and \textit{Mix}.
  \item \textbf{Reproducibility:} All intermediate JSONs (parsed schemas, diffs, votes, fusion schemas) are persisted to enable exact reruns and error analysis.
  \item \textbf{Failure handling:} Attributes whose voting confidence falls below a threshold are flagged for optional human review; otherwise, the pipeline proceeds autonomously.
\end{itemize}

Based on the above procedure, we obtain high-quality, automatically annotated raw data after filtering, providing a solid data foundation for subsequent tasks.

\subsubsection{VQA Auto-Generation Pipeline (VAGP)}

\begin{figure}[htbp] 
\centering
\includegraphics[width=\textwidth]{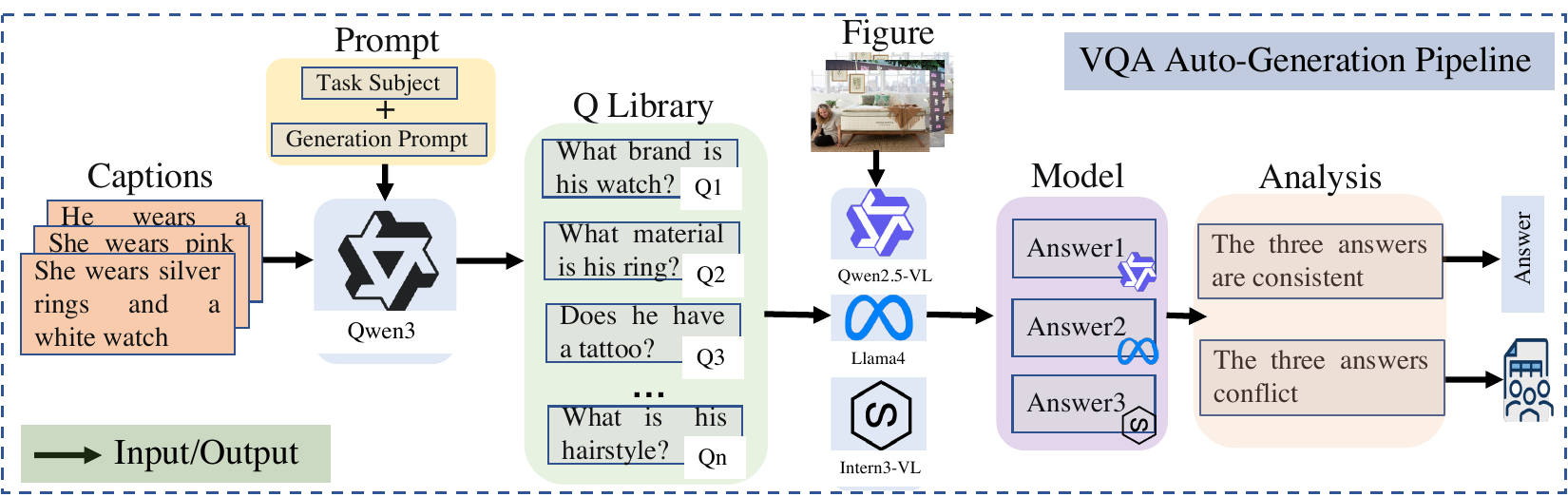}
\caption{Overall pipeline for \textbf{VAGP}, including three steps: VQA Generation, Vote and Manual Review.}
\label{fig:VAGP_pipline}
\end{figure}

We propose the VQA Auto-Generation Pipeline (VAGP), which exploits the high-quality raw data produced by the preceding automatic annotation process to automatically construct task-relevant VQA samples along three semantic dimensions: single-person, multi-person, and human--object. The goal is to generate questions and answers in a largely automated manner while ensuring strong relevance to the image and decidability from visible evidence; minimal human verification is invoked only when models disagree. The details can be found in Figure \ref{fig:VAGP_pipline}.

\paragraph{Stage 1: Question Proposal}
Given an image $I$, its high-quality caption, and a task topic with description, a general-purpose LLM is prompted to produce a set of candidate QAs. Prompts strictly require that questions be decidable from visible cues only, avoid subjective/causal reasoning and external knowledge, remove referential ambiguity, and obey length limits for both questions and answers. The generated questions explicitly cover the three dimensions (single-person / multi-person / human--object) and carry internal tags for type and expected answer format (e.g., counting, attribute, relation, action, spatial; boolean / numeric / short token).

\paragraph{Stage 2: Tri-MLLM Answering}
Each candidate question with the original image is fed in parallel to three heterogeneous or differently sized multimodal LLMs (MLLM1/2/3), yielding answers $\{A_{i,1}, A_{i,2}, A_{i,3}\}$. To enable robust comparison, answers are normalized by lowercasing, stripping punctuation, aligning numbers and number words, and mapping common synonyms. Format checks are enforced (booleans restricted to \texttt{yes/no}, counting to non-negative integers, colors/apparel/actions constrained by controlled vocabularies). Questions that show evident conflicts with the caption are discarded.

\paragraph{Stage 3: Consistency Filtering}
Consistency is decided from the three answers. Exact three-way agreement is marked as an \emph{easy case} and accepted as a high-confidence automatic sample. Otherwise, the item is labeled \emph{hard/ambiguous} and sent to a small human-review pool; optionally, a ``two-way agreement'' weak-consistency tag is recorded to support semi-automatic correction or retraining. To maintain distributional balance, quotas or weighted sampling are applied across the three semantic dimensions and question types.

\paragraph{Implementation and Outputs}
The LLM for question generation and the three answering MLLMs are decoupled for flexible configuration; batched parallel inference with cache reuse reduces cost. All intermediates and final artifacts (questions, raw and normalized answers from the three models, consistency labels, and metadata) are stored in JSON for reproducibility and auditing. Each finalized VQA item is recorded as \[
(I, \text{question}, \text{answer}, \text{topic}, \text{type}, \text{normalized\_answer}, A_{i,1}, A_{i,2}, A_{i,3}, \text{consistency\_label}, \text{metadata}).
\]
Easy cases are used directly for training/evaluation, while hard/ambiguous cases undergo human verification or semi-automatic correction. Through this procedure, we obtain the training set for supervised fine-tuning (SFT) as well as the final test set.

\subsection{Evaluation and Analysis}
\begin{figure}[htbp] 
\centering
\includegraphics[width=\textwidth]{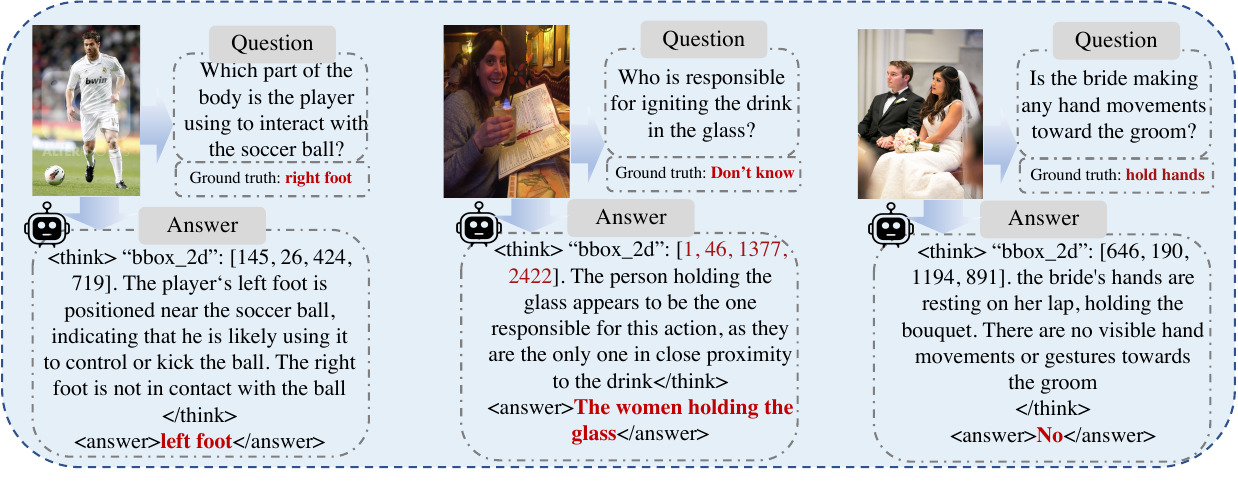}
\vspace{-0.2cm}
\caption{examples of bad cases, respectively covering spatial understanding, reasoning, and fine-grained perception.}
\label{fig:badcase_pipline}
\end{figure}

We first perform supervised fine-tuning (SFT) on the large-scale VQA data constructed along the three semantic dimensions, enabling the model to acquire basic domain knowledge and standardized output formats. However, this dataset contains many easy cases and only a few challenging ones, lacking targeted difficulty, and thus provides limited improvement to practical reasoning ability. To address this, we aim to curate a focused set of samples as a training set for reinforcement learning (RL). The detials can be found in Figure \ref{fig:badcase_pipline}.

Concretely, we evaluate an existing LVLM (e.g., Qwen2.5-VL-7B) on the test set, collect all incorrectly answered instances, and conduct a systematic bad case analysis to summarize common failure modes in reasoning. For these error cases, we instruct the model to produce a chain-of-thought along with the final answer, in order to precisely locate the source of errors. The corresponding prompt template is as follows:

\begin{tcolorbox}[
  enhanced,
  colback=gray!10,
  colframe=black,
  boxsep=4pt,
  left=8pt,right=8pt,top=8pt,bottom=8pt,
  arc=6pt,
  width=\linewidth
]
\itshape
This is a human-centric visual question answering task. Rely only on visible evidence from the image; avoid subjective assumptions and external knowledge. Before providing the answer, follow these steps: \textbf{Step 1.} Select and output the bounding box of the person or object most relevant to the question; \textbf{Step 2.} Based on the selected box, provide concise evidence bullet points and a brief, observable reasoning summary; \textbf{Step 3.} Give a concise answer in the format specified by the question.
\end{tcolorbox}

We analyzed all bad cases and identified three recurring capability gaps exhibited by current LVLMs on our benchmark: spatial relations, fine-grained perception, and calibration of reasoning depth. Detailed findings are as follows:
\begin{itemize}
  \item \textbf{Spatial relations.} The model struggles to determine whether a question is posed from the camera/viewer perspective or the agent's first-person perspective, and it frequently errs on absolute vs.\ relative spatial judgments. For example, an athlete is striking the ball with the right foot, so the correct answer is ``right foot.'' However, because the ball appears on the left side of the image, the model answers ``left foot.'' The chain-of-thought reveals the model failed to disambiguate the intended viewpoint.

  \item \textbf{Fine-grained perception.} The model has difficulty perceiving small, overlapping, or background details, and lacks robust fine-grained categorization. In one case, the bride and groom are holding hands, so the correct answer is ``A. Yes, she is holding his hand.'' A bouquet partially occludes the bride's hand, leading the model to misperceive the scene and output ``C. No, her hands are occupied with something else.''

  \item \textbf{Reasoning depth.} The model is poorly calibrated with respect to how much reasoning is warranted. Some questions require direct description or binary judgment; others require limited inference grounded in visible cues. In one example, a woman holds a glass with no additional action shown and no evidence that she ignited the drink; the appropriate answer is ``I don't know.'' The model, however, infers that she lit the drink merely because she is touching the glass, constituting over-reasoning beyond the evidence.
\end{itemize}

We have accordingly constructed a targeted VQA dataset focused on three core challenges: spatial relation reasoning, fine-grained perception, and calibration of reasoning depth. The samples deliberately cover complex scenarios and confounders, including viewpoint ambiguity (first-person vs. camera view), confusion between absolute and relative positions, occlusion and small-object detection, within-class fine categorization and background interference, as well as prompts that require “appropriate reasoning” rather than over-inference. Compared with general VQA corpora, this collection features a more targeted and diagnostic difficulty distribution, enabling systematic characterization and mitigation of model weaknesses along these dimensions.

For training, we plan to use this collection in a dedicated reinforcement learning (RL) stage, including but not limited to: enforcing consistency with visible evidence via the reward function, regularizing reasoning span and confidence, and shaping rewards for spatial consistency and fine-grained recognition. By optimizing on these high-value samples with RL, we aim to substantially improve the model’s stability in spatial reasoning, accuracy in fine-grained perception, and calibration of reasoning depth, thereby achieving more reliable and interpretable performance in complex real-world VQA scenarios.
\section{Experiment}

\definecolor{darkblue}{rgb}{0.0, 0.0, 0.55}

\subsection{Setup}

The base model used in our training framework is Qwen2.5-VL-7B \cite{bai2025qwen2}, a general-purpose multimodal instruction-tuned model that demonstrates strong performance in visual understanding and system prompt adherence. It is not specifically optimized for advanced reasoning through dedicated reinforcement learning, making it a suitable backbone for evaluating reasoning-oriented improvements. We fine-tune the model using supervised fine-tuning (SFT). During training, the vision encoder is frozen, and only the projection layer and the large language model are updated.

\subsection{Training parameters}

During the SFT process, the following parameters were utilized: A learning rate of $5 \times 10^{-5}$, a warm-up of 0.02, and a train batch size of 3. We apply a cutoff length of 8192 for generated responses, and then trained the model for 2 epoch to obtain \textbf{Qwen2.5-VL-7B-SFT}.

\subsection{Evaluation}
To comprehensively evaluate the performance of our model, Qwen2.5-VL-7B-SFT, a series of experiments were conducted across multiple subsets. Table 3 presents a detailed comparison of accuracy percentages for Qwen2.5-VL-7B-SFT, Qwen2.5-VL-7B-Instruct \cite{bai2025qwen2}, InternVL3-8B \cite{zhu2025internvl3}. The evaluation set comprises 10 subjects in total. Person and Object includes \textbf{Subject 1} (function and usage) and \textbf{Subject 5} (actions and interactions). Multiple Person covers \textbf{Subject 2} (interactions and behaviors), \textbf{Subject 3} (social relationships), and \textbf{Subject 4} (clothing and accessories). Single Person consists of \textbf{Subject 6} (physical appearance), \textbf{Subject 7} (clothing and accessories), \textbf{Subject 8} (actions and behaviors), \textbf{Subject 9} (identity and background), and \textbf{Subject 10} (distinctive attributes).



\begin{table*}[h]
\vspace{-0.3cm}
\newcommand{\tabincell}[2]{\begin{tabular}{@{}#1@{}}#2\end{tabular}}
\begin{center}
\caption{Performance (Accuracy \%) comparison of different models across multiple subsets. The best results are highlighted in bold. Qwen2.5-VL-7B-SFT is the model trained on a custom dataset that we constructed.}
\vspace{0.1cm}
\label{Table_Comparison}
\begin{threeparttable}
\fontsize{9}{12}\selectfont
\rmfamily
\setlength{\tabcolsep}{0.65mm}
\begin{tabular}{l|c|c|c|c|c|c|c|c|c|c|>{\columncolor{blue!8}}c}
\midrule
    Model & sub1 & sub2 & sub3 & sub4 & sub5 & sub6 & sub7 & sub8 & sub9 & sub10 & Average \\

    \midrule
    \rowcolor{pink!30}
    Qwen2.5-VL-7B-SFT & \textbf{96.92} & 78.10 & \textbf{90.66} & \textbf{82.41} & \textbf{69.08} & \textbf{78.39} & \textbf{81.94} & \textbf{62.32} & \textbf{87.50} & \textbf{86.67} & \textbf{81.40} \\ 
    Qwen2.5-VL-7B-Instruct & 96.92 & \textbf{80.95} & 90.66 & 78.70 & 67.63 & 74.37 & 74.84 & 55.07 & 87.50 & 86.67 & 79.33 \\ 
    InternVL3-8B & 96.92 & 80.48 & 90.66 & 78.70 & 68.12 & 70.35 & 74.37 & 54.84 & 81.97 & 80.00 & 77.64 \\ 
    \bottomrule
    
\end{tabular}
\end{threeparttable}
\end{center}
\label{tab:evaluation}
\end{table*}
\vspace{-0.2cm}

\textbf{Overall Performance:} As shown in Table \ref{Table_Comparison}, Qwen2.5-VL-7B-SFT achieved an average accuracy of 81.40\%, outperforming all other models tested. This demonstrates the effectiveness of training on the custom dataset we constructed, which emphasized fine-grained perception and reasoning capabilities. The model consistently outperformed its competitors across most subsets, showcasing its robustness and generalization abilities.

\textbf{Subset Specific Performance:} Qwen2.5-VL-7B-SFT achieved the highest accuracy in most subsets (e.g., sub1, sub3, sub4, sub5, sub6, sub7, and sub8, sub9, sub10), with key improvements observed in sub4 (82.41\%), sub5 (69.08\%), sub6 (78.39\%), sub7 (81.94\%) and sub8 (62.32\%). These results highlight the model's strength in addressing fine-grained reasoning and perception tasks.
Qwen2.5-VL-7B-Instruct achieved the second-best overall performance, with an average accuracy of 79.33\%. InternVL3-8B lagged behind Qwen2.5-VL-7B-SFT, achieving average accuracies of 77.64\%. These models struggled with subsets that required nuanced reasoning, such as sub5 and sub8.

\textbf{Key Observations:} Qwen2.5-VL-7B-SFT's strength in perception and reasoning: The results reflect the advantage of using our custom dataset, which was specifically designed to enhance fine-grained perception (e.g., appearance, pose, and parts) and reasoning (e.g., social relations, intent, and functionality). Performance gap in challenging subsets: While Qwen2.5-VL-7B-SFT excelled in most subsets, there is room for improvement in subsets like sub5 (69.08\%) and sub8 (62.32\%), where models faced challenges in handling intricate contextual cues and interactions.

In summary, the evaluation results strongly validate the effectiveness of our custom training dataset and methodology in enhancing the fine-grained perception and reasoning capabilities of Qwen2.5-VL-7B-SFT. The model's significant performance gains across multiple subsets establish it as a highly competitive approach for human-centric vision-language tasks.

\section{Discussions}

\textbf{Joint modeling of perception and reasoning is critical for performance gains.}  
Experimental results on MHPR show that jointly modeling fine-grained perception and high-level semantic reasoning is essential for improving human-centric visual understanding. Unlike datasets or training paradigms that focus on a single dimension (e.g., detection or attribute recognition), MHPR constrains models across the \emph{individual–multi-person–human–object} axes. This encourages models not only to perceive subtle visual cues (such as body parts, clothing, and pose), but also to reason about spatial relations, social interactions, and intent or functionality. This perception–reasoning integration largely explains why Qwen2.5-VL-7B achieves consistent improvements across multiple subsets.

\textbf{Format-aligned SFT data substantially improves stability and instruction following.}  
The experimental comparisons indicate that SFT-D, which is strictly aligned with the test set (T-D) in task composition and answer format, significantly enhances output stability and consistency. Qwen2.5-VL-7B-SFT outperforms Qwen2.5-VL-7B-Instruct and InternVL3-8B on most subsets, demonstrating that high-quality, format-consistent supervised data is often more effective than simply increasing model size for complex human-centric understanding tasks. This validates the role of SFT-D in MHPR as a stable alignment baseline.

\textbf{Bad-case–driven RL data effectively addresses core capability gaps.}  
Through systematic analysis of incorrect predictions on T-D, MHPR identifies three recurring failure modes in current LVLMs: spatial relations, fine-grained perception, and calibration of reasoning depth. Compared with generic VQA corpora, RL-D is explicitly constructed to target these weaknesses, covering viewpoint ambiguity (first-person vs.\ camera view), confusion between relative and absolute spatial judgments, occlusion and small-object perception, and cases where ``I don’t know'' is the appropriate answer. Such challenge-focused data provides high-value learning signals for reinforcement learning, improving robustness on difficult instances.

\textbf{Spatial understanding and viewpoint modeling remain major challenges.}  
Bad-case analysis reveals that errors related to spatial reasoning constitute a large portion of overall failures. Models frequently confuse image-based left/right with the agent’s own left/right, or fail to distinguish between camera and actor viewpoints. This suggests that scale alone is insufficient to resolve spatial reasoning issues, and that future work may benefit from explicitly modeling viewpoints or enforcing spatial-consistency constraints during training \cite{zhang2025open3dvqa, zhang2024countercurate, ai2025m2, ouyang2025spatial}.

\textbf{Fine-grained perception is highly sensitive to occlusion and background clutter.}  
Despite MHPR’s emphasis on fine-grained attributes, models remain vulnerable to occlusion, small targets, and complex backgrounds. For example, partial occlusion of hands by objects (e.g., a bouquet) often leads to incorrect judgments about hand interactions. These findings highlight the need for more robust local perception mechanisms, such as region-level supervision or localized reward shaping, to further enhance fine-grained recognition.

\textbf{Calibration of reasoning depth is more important than reasoning length.}  
We observe that models may either over-reason—drawing unwarranted conclusions without sufficient visual evidence—or under-calibrate by producing unnecessarily verbose reasoning for simple perceptual judgments. MHPR explicitly treats reasoning-depth calibration as a core challenge, emphasizing evidence-grounded and appropriately scoped reasoning. This suggests that high-quality human-centric understanding requires not only the ability to reason, but also the ability to determine \emph{when} and \emph{how much} reasoning is warranted \cite{wang2025perception, bigverdi2025perception}.

\textbf{Small models can be highly competitive with well-designed data and tasks.}  
Notably, Qwen2.5-VL-7B trained with MHPR achieves near-parity with, and in some cases outperforms, substantially larger models such as Qwen2.5-32B-Instruct. This result underscores the importance of carefully designed multidimensional benchmarks, targeted training objectives, and high-quality automated annotation pipelines. Together, these factors can significantly amplify the capabilities of relatively small models, offering more practical and cost-effective solutions for real-world human-centric vision–language applications.
\section{Related Works}

\subsection{Human-Centric Benchmark}
RefCOCO \cite{yu2016modeling} and RefCOCO+ \cite{yu2016modeling} are large-scale referring expression benchmarks built upon MSCOCO, where images are selected to contain multiple instances of the same object category and annotated with natural language expressions that uniquely identify a target object. RefCOCO allows unrestricted descriptions, while RefCOCO+ prohibits explicit location words to encourage appearance-based discrimination; both primarily evaluate language-guided visual grounding and object-level comprehension. However, these datasets focus on single-object localization and short, discriminative phrases, lacking fine-grained human-centric attributes, multi-person relational modeling, and higher-level reasoning (e.g., social relations, intent, and functionality), which limits their ability to assess joint perception–reasoning capabilities compared with MHPR.   HERM-Bench \cite{li2024herm} is a human-centric multimodal benchmark designed to evaluate MLLMs on fine-grained person understanding. It spans eight dimensions covering both basic perception (e.g., appearance, pose, grounding) and complex understanding (e.g., multi-person relations and reasoning), using multiple-choice and localization tasks. Unlike general benchmarks. Compared with MHPR, HERM-Bench mainly focuses on fine-grained human perception and basic relational understanding. Its reasoning is largely grounded in visible cues and structured QA formats. In contrast, MHPR emphasizes deeper semantic reasoning, such as social relations and intent inference. Therefore, HERM-Bench is relatively limited in evaluating higher-level joint perception–reasoning capabilities. HC-RefLoCo \cite{wei2024large} substantially extends earlier human-centric REC benchmarks (e.g., HC-RefCOCO/+/g) by introducing large-scale data and long, detailed referring expressions (avg. 93.2 words) with rich vocabulary and sentence-level subject labels, enabling fine-grained and scale-aware evaluation of multimodal grounding under diverse IoU criteria. Unlike prior benchmarks that rely on short phrases and limited samples, it emphasizes long-context understanding and precise localization. HC-RefLoCo primarily focuses on explicit visual–text alignment for long-context instance grounding, and thus lacks the ability to evaluate higher-level joint perception–reasoning over complex social relations, intentions, and implicit human semantics. In contrast, our MHPR explicitly addresses these limitations by incorporating tasks that require deeper relational understanding and human-centered reasoning beyond visible cues, enabling a more comprehensive assessment of advanced multimodal reasoning capabilities. HumanRef \cite{jiang2025referring} focuses on the task of referring to any person, requiring models to detect all individuals in an image that match a given natural language description. However, unlike our MHPR, HumanRef primarily emphasizes multi-instance detection, attribute recognition, spatial relations, interactions, and celebrity identification within explicit visual evidence. It does not evaluate deeper joint perception–reasoning over complex social dynamics, implicit intentions, or high-level human-centered semantics. In contrast, our MHPR explicitly addresses these limitations by introducing tasks that require advanced relational reasoning, implicit social understanding, and comprehensive human-centered perception beyond straightforward visual-text alignment. Face-Human \cite{qin2025face}  is designed to comprehensively evaluate multimodal assistants’ face and human understanding abilities through a hierarchical taxonomy covering perception and reasoning across facial attributes, expressions, identity, actions, spatial relations, and social relations.  Unlike our MHPR, it focuses on explicit recognition and shallow task-level reasoning in controlled settings, without evaluating deeper human-centered perception and implicit social reasoning. In contrast, MHPR targets these advanced capabilities.

\subsection{Reinforcement Learning for MLLMs}

Jiang et al. propose VLM-R3 \cite{jiang2025vlm}, a framework that enables MLLMs to dynamically ground and refine visual regions during reasoning through interleaved visual–textual chains. They introduce the VLIR dataset and a Region-Conditioned Reinforcement Policy Optimization (R-GRPO) strategy to train this capability, addressing the lack of dynamic visual grounding and fine-grained visual–text interaction in existing models.  Shen et al. propose VLM-R1 \cite{shen2025vlm}, an R1-style reinforcement learning framework that enhances VLMs with rule-based rewards and GRPO. By designing task-specific rewards for visual grounding and detection, it improves generalization and mitigates reward hacking compared to supervised fine-tuning. Zhang et al. propose R1-VL \cite{zhang2025r1}, a reinforcement learning framework based on GRPO that enhances MLLMs with dense step-wise reasoning rewards. By introducing step-wise reasoning accuracy and validity rewards to mitigate sparse outcome-level signals, it improves structured reasoning and overall performance beyond supervised fine-tuning and vanilla GRPO. Huang et al. propose Hint-GRPO \cite{huang2025boosting}, a text-debiased reinforcement learning framework that enhances MLLM reasoning under the GRPO paradigm. By introducing adaptive hint injection to improve data utilization and a test-time text-bias calibration mechanism to strengthen visual grounding, it mitigates low-reward inefficiency and text-only reliance in GRPO, achieving superior performance on complex multimodal reasoning tasks. Zhan et al. propose Vision-R1 \cite{zhan2025vision}, a vision-guided R1-style reinforcement learning framework that enhances LVLMs through a criteria-driven reward function and progressive rule refinement. By leveraging vision-specific feedback (e.g., format, recall, and precision rewards) without human preference data or reward models, it improves object localization performance and generalization beyond supervised fine-tuning. Chen et al. propose GRPO-CARE \cite{chen2025grpo}, a consistency-aware reinforcement learning framework that enhances MLLMs by jointly optimizing answer correctness and reasoning coherence. By introducing an adaptive, group-relative consistency bonus via reference-likelihood calibration and removing strict KL penalties, it mitigates reasoning–answer inconsistency in standard outcome-supervised GRPO and improves generalization and interpretability in multimodal video reasoning tasks. 

Existing multimodal reasoning approaches (e.g., large-scale SFT or reinforcement learning frameworks) primarily optimize for answer correctness and process consistency, yet lack structural modeling of key capabilities required in human-centric scenarios. As a result, they remain insufficient in the three aspects emphasized by MHPR. For Spatial Relations, the absence of explicit viewpoint and reference-frame modeling often leads to confusion between absolute and relative directions. For Fine-grained Perception, the reliance on global semantic representations makes models vulnerable to small objects, occlusion, and part-level details. For Reasoning Depth Calibration, although generating reasoning chains is encouraged, the lack of evidence-grounded constraints frequently results in over-inference or unwarranted confidence.

\section{Conclusion}
In this paper, we introduced MHPR (Multidimensional Human Perception and Reasoning), a comprehensive benchmark designed to address the critical gaps in human-centric perception and reasoning for large vision-language models (LVLMs). MHPR goes beyond existing benchmarks by evaluating models across three core scenario dimensions—individual, multi-person, and human–object interactions—encompassing fine-grained attributes (e.g., appearance, clothing, pose) and high-level semantics (e.g., social relations, spatial reasoning, intent, and functionality). This multidimensional evaluation framework enables a deeper and more holistic understanding of human-centric scenes.

To complement the benchmark, we proposed a multi-level data design (C-RD, SFT-D, RL-D, and T-D) and an automated caption/VQA generation pipeline (ACVG). These innovations ensure high-quality, scalable data generation while reducing manual efforts. Our pipeline employs techniques such as multi-model voting and consistency correction, resulting in precise annotations and robust task alignment.

Using MHPR, we trained Qwen2.5-VL-7B, yielding the fine-tuned Qwen2.5-VL-7B-SFT model. Extensive evaluations demonstrated that Qwen2.5-VL-7B-SFT outperforms other models, achieving an average accuracy of 81.88\% across multiple subsets. It consistently excelled in fine-grained perception and reasoning tasks, achieving near-parity with much larger models, validating the effectiveness of our custom dataset and training methodology.

\clearpage

\bibliography{iclr2025_conference}

@inproceedings{yu2016modeling,
  title={Modeling context in referring expressions},
  author={Yu, Licheng and Poirson, Patrick and Yang, Shan and Berg, Alexander C and Berg, Tamara L},
  booktitle={European conference on computer vision},
  pages={69--85},
  year={2016},
  organization={Springer}
}

@article{li2024herm,
  title={Herm: Benchmarking and enhancing multimodal llms for human-centric understanding},
  author={Li, Keliang and Yang, Zaifei and Zhao, Jiahe and Shen, Hongze and Hou, Ruibing and Chang, Hong and Shan, Shiguang and Chen, Xilin},
  journal={arXiv preprint arXiv:2410.06777},
  year={2024}
}

@article{wei2024large,
  title={A large-scale human-centric benchmark for referring expression comprehension in the LMM era},
  author={Wei, Fangyun and Zhao, Jinjing and Yan, Kun and Zhang, Hongyang and Xu, Chang},
  journal={Advances in Neural Information Processing Systems},
  volume={37},
  pages={69566--69587},
  year={2024}
}

@inproceedings{jiang2025referring,
  title={Referring to any person},
  author={Jiang, Qing and Wu, Lin and Zeng, Zhaoyang and Ren, Tianhe and Xiong, Yuda and Chen, Yihao and Qin, Liu and Zhang, Lei},
  booktitle={Proceedings of the IEEE/CVF International Conference on Computer Vision},
  pages={21667--21678},
  year={2025}
}

@article{qin2025face,
  title={Face-Human-Bench: A Comprehensive Benchmark of Face and Human Understanding for Multi-modal Assistants},
  author={Qin, Lixiong and Ou, Shilong and Zhang, Miaoxuan and Wei, Jiangning and Zhang, Yuhang and Song, Xiaoshuai and Liu, Yuchen and Wang, Mei and Xu, Weiran},
  journal={arXiv preprint arXiv:2501.01243},
  year={2025}
}

@article{jiang2025vlm,
  title={{VLM-R}$^{3}$: Region Recognition, Reasoning, and Refinement for Enhanced Multimodal Chain-of-Thought},
  author={Jiang, Chaoya and Heng, Yongrui and Ye, Wei and Yang, Han and Xu, Haiyang and Yan, Ming and Zhang, Ji and Huang, Fei and Zhang, Shikun},
  journal={arXiv preprint arXiv:2505.16192},
  year={2025}
}

@article{shen2025vlm,
  title={Vlm-r1: A stable and generalizable r1-style large vision-language model, 2025},
  author={Shen, Haozhan and Liu, Peng and Li, Jingcheng and Fang, Chunxin and Ma, Yibo and Liao, Jiajia and Shen, Qiaoli and Zhang, Zilun and Zhao, Kangjia and Zhang, Qianqian and others},
  journal={URL https://arxiv. org/abs/2504.07615},
  year={2025}
}

@article{zhang2025r1,
  title={R1-vl: Learning to reason with multimodal large language models via step-wise group relative policy optimization},
  author={Zhang, Jingyi and Huang, Jiaxing and Yao, Huanjin and Liu, Shunyu and Zhang, Xikun and Lu, Shijian and Tao, Dacheng},
  journal={arXiv preprint arXiv:2503.12937},
  year={2025}
}

@article{huang2025boosting,
  title={Boosting mllm reasoning with text-debiased hint-grpo},
  author={Huang, Qihan and Dai, Weilong and Liu, Jinlong and He, Wanggui and Jiang, Hao and Song, Mingli and Chen, Jingyuan and Yao, Chang and Song, Jie},
  journal={arXiv preprint arXiv:2503.23905},
  year={2025}
}

@article{zhan2025vision,
  title={Vision-r1: Evolving human-free alignment in large vision-language models via vision-guided reinforcement learning},
  author={Zhan, Yufei and Zhu, Yousong and Zheng, Shurong and Zhao, Hongyin and Yang, Fan and Tang, Ming and Wang, Jinqiao},
  journal={arXiv preprint arXiv:2503.18013},
  year={2025}
}

@article{chen2025grpo,
  title={GRPO-CARE: Consistency-Aware Reinforcement Learning for Multimodal Reasoning},
  author={Chen, Yi and Ge, Yuying and Wang, Rui and Ge, Yixiao and Cheng, Junhao and Shan, Ying and Liu, Xihui},
  journal={arXiv preprint arXiv:2506.16141},
  year={2025}
}

@inproceedings{fu2025video,
  title={Video-mme: The first-ever comprehensive evaluation benchmark of multi-modal llms in video analysis},
  author={Fu, Chaoyou and Dai, Yuhan and Luo, Yongdong and Li, Lei and Ren, Shuhuai and Zhang, Renrui and Wang, Zihan and Zhou, Chenyu and Shen, Yunhang and Zhang, Mengdan and others},
  booktitle={Proceedings of the Computer Vision and Pattern Recognition Conference},
  pages={24108--24118},
  year={2025}
}

@inproceedings{li2024mvbench,
  title={Mvbench: A comprehensive multi-modal video understanding benchmark},
  author={Li, Kunchang and Wang, Yali and He, Yinan and Li, Yizhuo and Wang, Yi and Liu, Yi and Wang, Zun and Xu, Jilan and Chen, Guo and Luo, Ping and others},
  booktitle={Proceedings of the IEEE/CVF Conference on Computer Vision and Pattern Recognition},
  pages={22195--22206},
  year={2024}
}

@article{zhouhumanvbench,
  title={HumanVBench: Probing Human-Centric Video Understanding in MLLMs with Automatically Synthesized Benchmarks},
  author={Zhou, Ting and Chen, Daoyuan and Jiao, Qirui and Ding, Bolin and Li, Yaliang and Shen, Ying}
}

@inproceedings{dong2025moga,
  title={MoGA: 3D generative avatar prior for monocular gaussian avatar reconstruction},
  author={Dong, Zijian and Duan, Longteng and Song, Jie and Black, Michael J and Geiger, Andreas},
  booktitle={Proceedings of the IEEE/CVF International Conference on Computer Vision},
  pages={13304--13314},
  year={2025}
}

@article{chen2025socialnav,
  title={SocialNav: Training Human-Inspired Foundation Model for Socially-Aware Embodied Navigation},
  author={Chen, Ziyi and Guo, Yingnan and Chu, Zedong and Luo, Minghua and Shen, Yanfen and Sun, Mingchao and Hu, Junjun and Xie, Shichao and Yang, Kuan and Shi, Pei and others},
  journal={arXiv preprint arXiv:2511.21135},
  year={2025}
}

@article{dai2025humanvlm,
  title={Humanvlm: Foundation for human-scene vision-language model},
  author={Dai, Dawei and Xu, Long and Li, Yutang and Zhang, Yuanhui and Xia, Shuyin},
  journal={Information Fusion},
  pages={103271},
  year={2025},
  publisher={Elsevier}
}

@misc{lu2023delvingdeeperdatascaling,
      title={Delving Deeper into Data Scaling in Masked Image Modeling}, 
      author={Cheng-Ze Lu and Xiaojie Jin and Qibin Hou and Jun Hao Liew and Ming-Ming Cheng and Jiashi Feng},
      year={2023},
      eprint={2305.15248},
      archivePrefix={arXiv},
      primaryClass={cs.CV},
      url={https://arxiv.org/abs/2305.15248}, 
}

@article{bai2025qwen2,
  title={Qwen2. 5-vl technical report},
  author={Bai, Shuai and Chen, Keqin and Liu, Xuejing and Wang, Jialin and Ge, Wenbin and Song, Sibo and Dang, Kai and Wang, Peng and Wang, Shijie and Tang, Jun and others},
  journal={arXiv preprint arXiv:2502.13923},
  year={2025}
}

@article{zhu2025internvl3,
  title={Internvl3: Exploring advanced training and test-time recipes for open-source multimodal models},
  author={Zhu, Jinguo and Wang, Weiyun and Chen, Zhe and Liu, Zhaoyang and Ye, Shenglong and Gu, Lixin and Tian, Hao and Duan, Yuchen and Su, Weijie and Shao, Jie and others},
  journal={arXiv preprint arXiv:2504.10479},
  year={2025}
}

@article{zhang2025open3dvqa,
  title={Open3dvqa: A benchmark for comprehensive spatial reasoning with multimodal large language model in open space},
  author={Zhang, Weichen and Zhou, Zile and Zheng, Zhiheng and Gao, Chen and Cui, Jinqiang and Li, Yong and Chen, Xinlei and Zhang, Xiao-Ping},
  journal={arXiv preprint arXiv:2503.11094},
  year={2025}
}

@article{zhang2024countercurate,
  title={Countercurate: Enhancing physical and semantic visio-linguistic compositional reasoning via counterfactual examples},
  author={Zhang, Jianrui and Cai, Mu and Xie, Tengyang and Lee, Yong Jae},
  journal={arXiv preprint arXiv:2402.13254},
  year={2024}
}

@article{wang2025perception,
  title={Perception-aware policy optimization for multimodal reasoning},
  author={Wang, Zhenhailong and Guo, Xuehang and Stoica, Sofia and Xu, Haiyang and Wang, Hongru and Ha, Hyeonjeong and Chen, Xiusi and Chen, Yangyi and Yan, Ming and Huang, Fei and others},
  journal={arXiv preprint arXiv:2507.06448},
  year={2025}
}

@article{ai2025m2,
  title={M2-reasoning: Empowering mllms with unified general and spatial reasoning},
  author={AI, Inclusion and Wang, Fudong and Liu, Jiajia and Chen, Jingdong and Zhou, Jun and Ji, Kaixiang and Ru, Lixiang and Guo, Qingpei and Zheng, Ruobing and Li, Tianqi and others},
  journal={arXiv preprint arXiv:2507.08306},
  year={2025}
}

@inproceedings{bigverdi2025perception,
  title={Perception tokens enhance visual reasoning in multimodal language models},
  author={Bigverdi, Mahtab and Luo, Zelun and Hsieh, Cheng-Yu and Shen, Ethan and Chen, Dongping and Shapiro, Linda G and Krishna, Ranjay},
  booktitle={Proceedings of the Computer Vision and Pattern Recognition Conference},
  pages={3836--3845},
  year={2025}
}

@article{ouyang2025spatial,
  title={Spatial-r1: Enhancing mllms in video spatial reasoning},
  author={Ouyang, Kun},
  journal={arXiv e-prints},
  pages={arXiv--2504},
  year={2025}
}
\bibliographystyle{iclr2025_conference}
\clearpage
\appendix
\section{Single Person}

\newtcolorbox{promptbox}[1]{
  breakable,
  colback=white,
  colframe=black,
  boxrule=0.8pt,
  arc=0pt,
  left=6pt,
  right=6pt,
  top=6pt,
  bottom=6pt,
  enhanced,
  title=#1,
  coltitle=white,
  colbacktitle=black,
  fonttitle=\bfseries\centering, 
  boxed title style={
    sharp corners,
    boxrule=0pt,
    halign=center             
  }
}

\begin{promptbox}{Single-Person Description Prompt}
{\large\bfseries\color{red}Prompt:}
You are an expert in the field of visual description, and I will provide you with an image of a person. Please detail the various aspects of the person depicted in the image based on the following guidelines: First, describe the person's basic information, including gender, age group, and ethnic characteristics. Next, carefully depict the person's appearance. Describe their hairstyle, noting the length, color, and style. Then, focus on facial features, providing detailed descriptions of the characteristics such as the size and shape of the eyes, the height of the nose bridge, and the thickness of the lips. Continue to describe the facial contour, skin color (such as fair or dark), and condition, while paying attention to facial expressions, such as smiling, seriousness, or surprise. Subsequently, describe the person's body features, including height, body type, and posture. Observe their style of dress, including the type of clothing, color combinations, materials, and the style and type of shoes. Provide detailed descriptions of any accessories, such as jewelry, glasses, hats, or bags, noting their details and colors. Next, observe the person's actions and posture. Describe their overall stance and hand movements, as well as the emotional state conveyed by their body language. Describe the person's specific actions and infer their possible occupational characteristics or social roles. Continue by describing the background environment and atmosphere, noting whether the setting is indoors or outdoors, and infer the season or weather through clothing or other details. Describe the emotional and cultural atmosphere of the scene, such as whether there are traditional costumes or modern architecture. Finally, pay attention to any special markings and personalized features. Include the location and pattern of tattoos, the location and shape of scars or birthmarks, and any special makeup or habitual body language. Ensure your description is coherent and vivid, enabling listeners to form a clear visual impression without seeing the image. Use complete sentences to present your description, rather than providing information in a list format. If certain aspects are not visible or are obscured in the image, do not describe them, and refrain from making assumptions.

\vspace{0.2cm}

{\large\bfseries\color{red}Example of a generated caption:} In the image, we see a man standing on a red carpet, likely at a formal event such as a film festival or awards ceremony. His gender is male, and he appears to be in his late twenties or early thirties. Based on his facial features, he seems to belong to an East Asian ethnic group. The man has short, dark hair that is neatly styled with a slight side parting, giving him a polished and professional appearance. His hairstyle is well-groomed, adding to his overall sophisticated look. His eyes are of medium size, almond-shaped, and convey a sense of warmth and friendliness. The bridge of his nose is straight and well-defined, contributing to his balanced facial structure. His lips are of average thickness, and he has a pleasant smile that reaches his eyes, suggesting a friendly and approachable demeanor. His skin tone is fair, and his complexion appears smooth and healthy. He is wearing a subtle makeup that enhances his natural features without being overly noticeable. Regarding his body features, the man appears to be of average height with a slim yet athletic build. His posture is upright and confident, indicating poise and self-assurance. He is dressed in a classic black velvet tuxedo, which fits him well and accentuates his slender figure. The tuxedo is paired with a crisp white dress shirt and a black bow tie, creating a timeless and elegant ensemble. His shoes are not visible. He is holding a plush toy in one hand, which adds a touch of whimsy and charm to his otherwise formal attire. There are no other accessories visible in the image, such as jewelry, glasses, hats, or bags. The man's actions and posture suggest that he is comfortable and at ease in the spotlight. He is waving with one hand, a gesture that conveys a sense of greeting and acknowledgment towards the audience or photographers. His overall stance is relaxed yet composed, indicating that he is accustomed to public appearances. This behavior, combined with his attire and the setting, suggests that he may be a public figure, possibly an actor, musician, or another type of celebrity who frequently attends high-profile events. The background environment is bustling with activity, typical of a red carpet event. There are numerous people in the background, some of whom appear to be photographers and journalists capturing the moment. The red carpet itself is vibrant and serves as a striking contrast to the man's dark suit. The atmosphere is lively and glamorous, with a sense of excitement and anticipation in the air. The lighting is bright, highlighting the man and the red carpet, while the crowd in the background is slightly blurred, drawing attention to the main subject. There are no visible special markings or personalized features such as tattoos, scars, or birthmarks on the man in this image. His makeup is natural and understated, enhancing his features without altering them significantly. His body language is open and engaging, further reinforcing the impression of a person who is comfortable in the public eye and enjoys interacting with others. Overall, the image portrays a charismatic and stylish individual who is clearly at home in a high-profile social setting.
\end{promptbox}

\section{Multi-Person}

\begin{promptbox}{Multi-Person Description Prompt}

{\large\bfseries\color{red}Prompt:}
You are an expert in image detail analysis. I will provide you with an image and ask you to analyze the relationship characteristics between characters in the image. More specifically, it involves the characteristics of person relationships between two or more people. Please describe this image from the following perspectives, including but not limited to:

\begin{itemize}
  \item \textbf{Spatial dimension:}  
  Considering the positions of individuals, it may be necessary to analyze their relative positions, arrangement relationships, and distance; It may be necessary to reconsider the orientation and posture of the human body; From the perspective of spatial arrangement, it may be necessary to analyze the position occupying the center, marginal positions, height differences, etc; Considering spatial, background, and environmental cues, there may be factors such as whether the space is shared or divided, whether there are clear task areas, and environmental arrangements.

  \item \textbf{Interaction and behavioral dimensions:}  
  From the perspective of physical contact, one can consider whether there is physical contact, close proximity, and corresponding movements; From the perspective of eye contact and facial expressions, one can consider whether there is eye contact, facial expressions, and similar emotions; In terms of verbal or nonverbal communication, including conversational behavior, gesture expression, and interaction between postures and postures; In terms of role interaction, there may be leading and obeying, providing and receiving help, confronting conflicts, collaborating and cooperating, companionship and leisure, etc; There are also situational interactions, including specific occasion interactions, identity cues, spatial layout, etc.

  \item \textbf{Social relationship dimension:}  
  Family or kinship relationships, as well as friendship relationships, can be considered. There are also workplace relationships, etc; In terms of social roles, there are some relationships between professional roles and their service recipients.

  \item \textbf{Clothing and Accessories:}  
  From the perspective of clothing characteristics, there are uniformity and similarity in clothing appearance styles, hierarchical differences reflected in professional or high-end clothing, matching between clothing, cultural or ethnic clothing, seasonal clothing, brand or luxury brand logos; From the perspective of accessories, there are accessories consistency, identity symbol accessories, functional accessories, decorative accessories, and interactive accessories.

  \item \textbf{Environment and Scene:}  
  The scene types include but are not limited to indoor scenes, outdoor scenes, and special event scenes; Environmental elements include but are not limited to iconic objects in the background, weather and lighting, and time; In terms of the interaction between people and the environment, whether multiple people participate in the same activity, whether there are guiding actions, etc.
\end{itemize}

The above is a list of possible descriptive perspectives for you. You can describe it from other perspectives based on specific image features. When describing, you need to meet the following requirements:

\begin{itemize}
  \item You need to selectively describe this image from the above dimensions based on the theme of the current image.
  \item You need to describe this image in a fluent paragraph, without listing or explicitly stating a certain dimension.
  \item For crowded scenarios, you need to refer clearly to the person you are describing based on their distinctive features.
  \item Just give affirmative content, and try not to say speculative or possible parts.
  \item What needs to be accurately distinguished is the left-right relationship. When using, you must clarify two methods of description. The first type is on the left or right side of the image, and the second type is on the left or right side of the person. You can't mix them together.
\end{itemize}

\end{promptbox}

\section{Person and Object}
\begin{promptbox}{Person-Object Relationship Description Prompt}
{\large\bfseries\color{red}Prompt:}
You are an expert in image details, and I now need to describe the relationship between the specified person and related items in the image. Please provide a detailed description. The following are the attributes and categories that I focus on:

\begin{itemize}
  \item \textbf{The spatial relationship between people and objects:}  
  more specifically, it includes categories such as proximity relationship, containment relationship, hierarchical relationship, symmetry relationship, contrast relationship, directional relationship, distance relationship, up-down relationship, and surrounding relationship.

  \item \textbf{The action and interaction relationship between people and objects:}  
  including categories such as possession, use, observation, manipulation, support, interaction, transmission, adjustment, destruction, repair, creation, etc.

  \item \textbf{The emotional and motor relationships between people and objects:}  
  include categories such as love and cherish, curiosity and exploration, focus and engagement, dissatisfaction and boredom, nostalgia and memories, loss and regret, excitement and happiness, anxiety and tension, intimacy and dependence.

  \item \textbf{The relationship between the functions and uses of people and objects:}  
  including tool use, information acquisition, entertainment and leisure, learning and education, transportation and mobility, storage and organization, protection and safety, cooking and eating, cleaning and hygiene, decoration and beautification, communication and exchange, health and fitness, and other categories.

  \item \textbf{The relationship between people and objects in terms of environment and background:}  
  including cultural background, historical background, geographical environment, socio-economic background, technological background, ecology and environment, legal regulation, fashion and trends, seasons and climate, functional environment.

  \item \textbf{Cultural and social background relationship between people and objects:}  
  symbolic meaning, identity, traditional customs, aesthetic values, social norms, consumer culture, language and symbols, festivals, historical inheritance, etc.

  \item \textbf{The visual element background relationship between people and objects:}  
  color, shape, texture, contrast, proportion and scale, symmetry, pattern decoration, font layout, light and shadow effects, dynamic and static design, etc.
\end{itemize}

Here are some requirements:

\begin{itemize}
  \item I will provide you with the coordinates of the person and mark them with a red rectangular box in the picture. You only need to describe this person.
  \item You need to describe items that are related to the current target person, not unrelated items. Do not describe clothes, shoes, or other clothing. Do not describe headgear, scarves, or any other accessories. Do not describe glasses. Do not mention red boxes in the generated description. If there are no items related to the target person, simply answer ``there is no related items''.
  \item You need to describe it in fluent language, without listing attribute names. Do not mention the name of the relationship type.
  \item If a person in the picture does not have any related items, you must answer ``there is no related items''.
  \item Do not describe anything other than the item.
  \item Don't mention anyone else.
\end{itemize}

The coordinates of the target character in the picture are:

\end{promptbox}

\end{document}